\pdfoutput=1
\documentclass[11pt]{article}
\usepackage[final]{acl}
\usepackage{times}
\usepackage{latexsym}
\usepackage[T1]{fontenc}
\usepackage[utf8]{inputenc}
\usepackage{microtype}
\usepackage{inconsolata}
\usepackage{graphicx}
\usepackage{amsmath}
\usepackage{amssymb}
\usepackage{mathtools}
\usepackage{booktabs}
\usepackage{amssymb}
\usepackage{multicol}
\usepackage{booktabs}
\usepackage{multirow}
\usepackage{colortbl}
\usepackage{xcolor}
\usepackage{algorithm}
\usepackage{algpseudocode}
\usepackage{algorithmicx}
\usepackage{array}
\usepackage{subcaption}
\usepackage{tabularx}
\usepackage{colortbl}
\usepackage{makecell}
\usepackage{placeins}
\usepackage{float}
\usepackage{tikz,lipsum}
\usepackage[most]{tcolorbox}
\usepackage{natbib}
\usepackage{pgf} 
\newcommand{\dataset}{\textsc{MCR‐Bench}}

\newenvironment{packeditemize}{
\begin{list}{$\bullet$}{
\setlength{\labelwidth}{6pt}
\setlength{\itemsep}{0pt}
\setlength{\leftmargin}{\labelwidth}
\addtolength{\leftmargin}{\labelsep}
\setlength{\parindent}{0pt}
\setlength{\listparindent}{\parindent}
\setlength{\parsep}{0pt}
\setlength{\topsep}{3pt}}}{\end{list}}

\definecolor{white}{RGB}{253,253,253}
\definecolor{verylightgreen}{RGB}{242,250,244} 
\definecolor{lightgreen}{RGB}{226,243,232}
\definecolor{mediumgreen}{RGB}{184,226,197} 
\definecolor{green}{RGB}{142,208,160}
\definecolor{darkgreen}{RGB}{99,190,123}
\newcommand{\shadecell}[1]{%
    \ifdim #1pt > 150pt \cellcolor{darkgreen}#1%
    \else\ifdim #1pt > 100pt \cellcolor{green}#1%
    \else\ifdim #1pt > 75pt \cellcolor{mediumgreen}#1%
    \else\ifdim #1pt > 50pt \cellcolor{lightgreen}#1%
    \else\ifdim #1pt > 0pt \cellcolor{verylightgreen}#1%
    \else\cellcolor{white}#1%
    \fi\fi\fi\fi\fi}

\title{When Audio and Text Disagree: Benchmarking Text Bias in Large Audio-Language Models under Cross-Modal Inconsistencies}

\author{
Cheng Wang$^\dagger$ \quad 
Gelei Deng$^\ddagger$\thanks{* Corresponding Author} \quad 
Xianglin Yang$^\dagger$ \quad 
Han Qiu$^\S$ \quad 
Tianwei Zhang$^\ddagger$
\\
$^\dagger$ National University of Singapore \\
$^\ddagger$ Nanyang Technological University \\
$^\S$ Tsinghua University\\
\texttt{wangcheng@u.nus.edu} 
}

\begin{document}
\maketitle
\begin{abstract}
Large Audio-Language Models (LALMs) are enhanced with audio perception capabilities, enabling them to effectively process and understand multimodal inputs that combine audio and text. However, their performance in handling conflicting information between audio and text modalities remains largely unexamined. This paper introduces \dataset{}, the first comprehensive benchmark specifically designed to evaluate how LALMs prioritize information when presented with inconsistent audio-text pairs. Through extensive evaluation across diverse audio understanding tasks, we reveal a concerning phenomenon: when inconsistencies exist between modalities, LALMs display a significant bias toward textual input, frequently disregarding audio evidence. This tendency leads to substantial performance degradation in audio-centric tasks and raises important reliability concerns for real-world applications. We further investigate the influencing factors of text bias, and explore mitigation strategies through supervised finetuning, and analyze model confidence patterns that reveal persistent overconfidence even with contradictory inputs. These findings underscore the need for improved modality balance during training and more sophisticated fusion mechanisms to enhance the robustness when handling conflicting multi-modal inputs\footnote{The project is available at \url{https://github.com/WangCheng0116/MCR-BENCH}}.
\end{abstract}

\section{Introduction}

With the rise of Large Audio-Language Models (LALMs)~\citep{qwen-2-audio,salmonn,LTU-AS}, there has been significant progress in developing applications and systems capable of processing both auditory and textual information for complex tasks. These models, often built upon Large Language Models (LLMs) with specialized audio encoders, have demonstrated impressive capabilities in various audio-centric tasks including Audio Question Answering~\citep{ClothoAQA}, Sound Event Detection~\citep{Mesaros_2021}, and Speech Recognition~\citep{whisper}. The wide deployment of LALMs across various domains reflects their growing importance in bridging human auditory experience with machine intelligence.

\begin{figure}
    \centering
\includegraphics[width=0.9\linewidth]{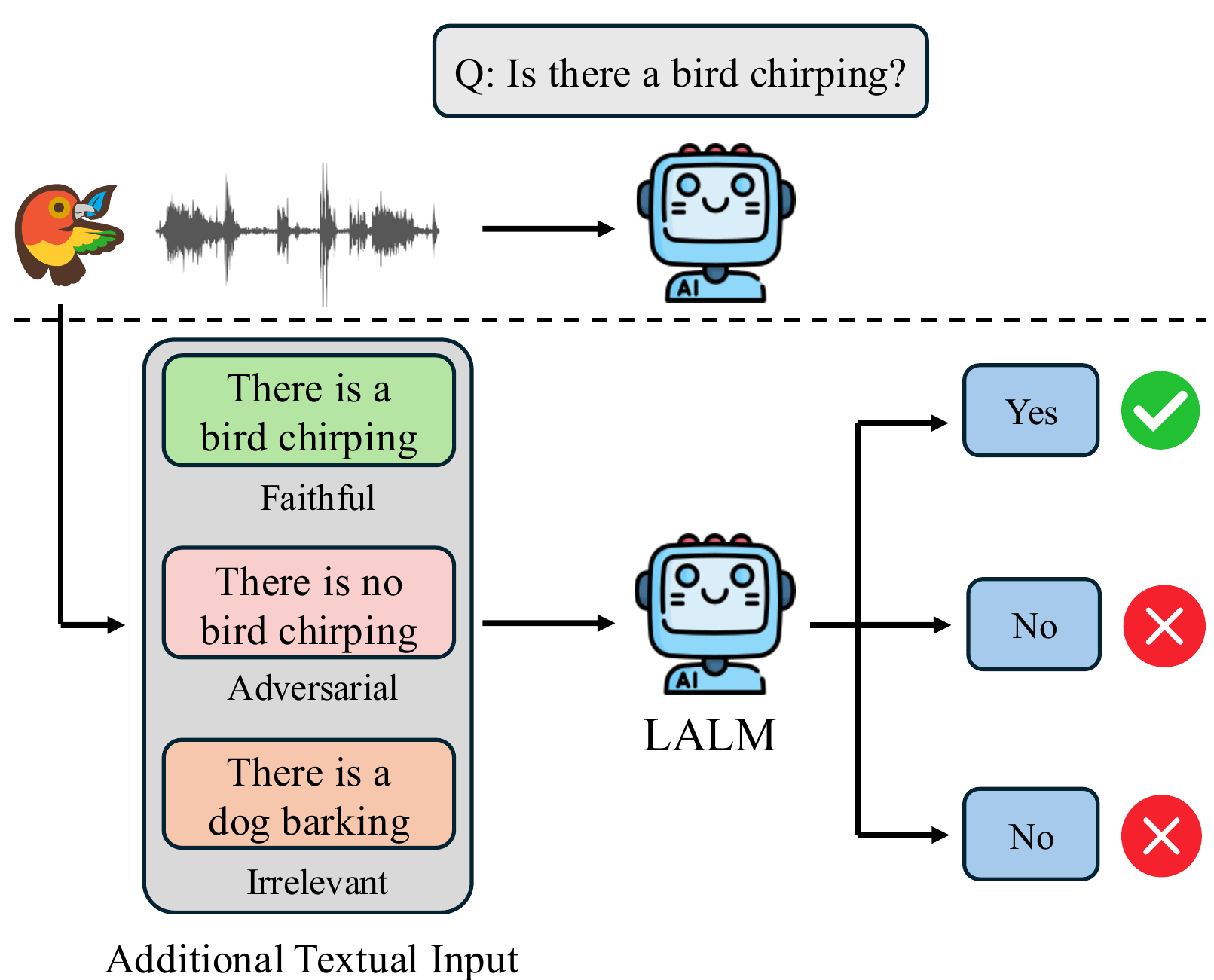}
    \caption{Illustration of LALMs handling users' input with conflicts across the text and audio modalities.}
    \vspace{-10pt}
    \label{fig:illustration}
\end{figure}

To facilitate the development of LALMs, numerous benchmarks and datasets have been established for performance evaluations \cite{wang2024audiobench,yang2024air}. However, they typically assume harmonious or complementary relationships between audio and text inputs. In particular, standard datasets often pair audio samples with accurate textual descriptions or questions that precisely align with the audio content. This idealized evaluation approach, while useful for basic capability assessment, fails to capture the \textit{robustness} of these models in handling real-world scenarios where the input of different modalities contains conflicting information. Researchers have demonstrated that the inconsistent inputs could significantly degrade the performance of LLMs~\citep{shi2023large, liu2024lost} or Large Vision-Language Models (LVLMs)~\citep{liu2025robustness, ailindeng}. However, there is still a lack of systematic investigation into how LALMs behave when faced with contradictory inputs, representing a significant gap in our understanding of these models' reliability.

The above research gap drives the motivation of our study, where we aim to systematically evaluate and mitigate the limitations of contemporary LALMs under conflicting modal information. We believe this is crucial for ensuring their safe and dependable use in real-world applications. We hypothesize that when faced with inconsistent audio and text inputs, LALMs may exhibit a bias toward one modality—either audio or text—over the other, potentially leading to suboptimal performance in audio-centric tasks. This preferential behavior could undermine the models' ability to effectively integrate and reconcile multi-modal data, which is essential for their robustness in complex, dynamic environments. 

To validate our hypothesis, we introduce \dataset{}, a comprehensive \underline{M}odal \underline{C}onflict \underline{R}esolution \underline{Bench}mark for LALMs. Departing from traditional clean audio-text pairs, \dataset{} comprises 3,000 specially constructed samples across three audio-centric tasks, where each audio input is systematically paired with adversarial, faithful, and irrelevant textual descriptions. Through extensive experiments evaluating six state-of-the-art LALMs on \dataset{}, we reveal a consistent and substantial preference for textual input over audio, leading to severe performance degradation in the presence of misleading text. This modality bias is evident across diverse tasks and model architectures, indicating a widespread issue in current LALM designs.

Beyond characterizing textual bias, we further explore mitigation strategies and analyze internal LLM state differences when processing clean versus contradictory samples.
We find that simple prompting techniques—such as bias-aware or audio-prioritized instructions—yield only limited improvements, while supervised finetuning on conflict-rich data offers more promising, though still incomplete, mitigation. Further analysis of model behavior reveals that LALMs remain highly confident even when relying on contradictory textual information, and internal representation studies suggest they internally detect cross-modal inconsistencies without appropriately modulating their outputs. These findings underscore a disconnect between latent awareness and output reasoning, highlighting the need for architectural and training-level innovations to achieve truly robust multi-modal reasoning in audio-language models.
Importantly, these insights illuminate a promising path forward: leveraging mechanism interpretation to develop new solutions for robust audio-language models.

\section{Related Work}
\textbf{LALMs Performance Benchmarking.}
Large Audio-Language Models (LALMs) have recently gained significant attention for their ability to process audio inputs and generate textual responses. Researchers have established task-specific benchmarks for audio understanding capabilities, such as AudioBench \citep{wang2024audiobench} and AIR-Bench \citep{yang2024air}. These benchmarks predominantly assume aligned or complementary audio-text relationships, leaving the models' behavior under conditions of modal conflict largely unexplored. While recent work has begun addressing evaluation comprehensiveness, the assumption of modal harmony persists, creating a critical gap in our understanding of LALMs' reliability in real-world scenarios where inputs across modalities may contain inconsistencies.

\noindent\textbf{Robustness of LALMs.}
Prior research has focused on two primary dimensions of audio models' robustness: vulnerability to adversarial attacks and resilience against natural perturbations. While \citet{carlini2018audio} and \citet{qin2019imperceptible} demonstrated concerning susceptibilities to targeted and imperceptible adversarial examples, defensive strategies such as data augmentation techniques proposed by \citet{park2019specaugment} and self-supervised learning frameworks from \citet{baevski2020wav2vec} have shown promise in improving model resilience. Despite these advances, the field requires more systematic evaluations and comprehensive frameworks to address the multifaceted challenges of real-world audio processing.

\noindent\textbf{Distraction in Inputs.}
Recent studies highlight the challenge of distraction in input processing across both language and multimodal models. For LLMs, \citet{huang2025breakingfocuscontextualdistraction} introduced Contextual Distraction Vulnerability, demonstrating how irrelevant but semantically coherent context significantly degrades model performance. To address this challenge, retrieval-augmented contrastive learning approaches have been explored to enhance focus on relevant information in long-context tasks \citep{wu2024reducingdistractionlongcontextlanguage}. The distraction problem extends to multimodal systems as well, with \citet{ailindeng} and \citet{liu2025robustness} systematically analyzing how Vision-Language Models exhibit substantial performance degradation when confronted with conflicting visual and textual inputs. These studies establish that inconsistent or distracting information across modalities presents a fundamental challenge for robust AI systems. Our work extends this line of inquiry to the audio domain, investigating how LALMs prioritize information when faced with similar cross-modal inconsistencies.

\section{\dataset{}}
We introduce \dataset{}, a benchmark specifically designed to evaluate how LALMs process and reconcile conflicting audio-text inputs. For each audio sample in our benchmark, we systematically construct three types of textual contexts:

\begin{packeditemize}
    \item \textbf{Faithful:} Accurate descriptions that correctly represent the audio content.
    \item \textbf{Adversarial:} Deliberately misleading descriptions that contradict the audio content.
    \item \textbf{Irrelevant:} Semantically unrelated descriptions that have minimal topical overlap with the audio content.
\end{packeditemize}

These variations allow us to systematically evaluate LALMs' ability to prioritize relevant audio information, resist misleading textual cues, and maintain robust performance when faced with conflicting or irrelevant cross-modal inputs. Below we elaborate how these three types of text variations are constructed. 

\subsection{Data Sources}
\dataset{} covers three different types of audio understanding tasks (sound question answering, speech emotion recognition, and vocal sound classification) to ensure a comprehensive evaluation across diverse audio domains. It is extensible for supporting other audio-text tasks as well.

\begin{packeditemize}
    \item \textbf{Audio Question Answering (AQA):} We utilize ClothoAQA~\citep{ClothoAQA}, a dataset comprising 1,991 audio samples from the Clotho~\citep{clotho} dataset, each paired with six crowdsourced questions and corresponding answers, totaling 35,838 question-answer pairs. This component evaluates natural language understanding of general audio content.
    
    \item \textbf{Speech Emotion Recognition (SER):} We incorporate MELD~\citep{meld}, a multimodal multi-party dataset containing over 1,400 dialogues and 13,000 utterances from the TV series Friends, annotated with seven emotion labels and sentiment. This tests the model performance on human speech with emotional content.
    
    \item \textbf{Vocal Sound Classification (VSCn):} We include VocalSound~\citep{VocalSound}, which features non-verbal human vocalizations across different acoustic conditions, challenging models to recognize human vocal sounds beyond speech.
\end{packeditemize}

\subsection{Text Variation Construction}
\label{subsec:text-variation}
To generate systematic variations in textual contexts, we create three distinct textual conditions for each audio sample:

\begin{packeditemize}
    \item \textbf{Faithful Text Generation:} We employ GPT-4o~\citep{gpt4o} with one-shot learning to create factual statements that accurately represent the audio content based on original question-answer pairs.
    
    \item \textbf{Adversarial Text Generation:} Using the same GPT-4o framework, we generate non-factual statements that directly contradict the audio content. Appendix~\ref{app:faithful-adv-prompt} shows the prompt template used for this adversarial generation process.
    
    \item \textbf{Irrelevant Text Selection:} We select irrelevant textual descriptions based on sentence similarity calculations between the true caption and all captions from AudioCaps~\citep{audiocaps}. We choose descriptions with minimal semantic overlap while maintaining plausible text structure.
\end{packeditemize}

\subsection{Evaluation Metrics}
To quantify modal conflict resolution capabilities of different LALMs, we define $N$ as the total number of samples, $C_{\text{neutral}}$ as the number of correct predictions under neutral conditions (where only audio input is provided without any textual description), and $C_{t}$ as the number of correct predictions with text condition $t \in \{{\textit{neu}}, {\textit{fth}}, {\textit{adv}}, {\textit{irr}}\}$ for faithful, adversarial, and irrelevant conditions respectively. For evaluation, we use a prompt template shown in Figure~\ref{fig:prompt_eval} that instructs models to answer questions while being aware that the provided textual descriptions may contain inaccuracies. Specifically, we use the following metrics.

\begin{figure}[hb]
    \centering
\begin{tcolorbox}[fonttitle=\bfseries,title=Prompts for Evaluation, fontlower=\small]
The following text provides descriptions derived from audio content. This information may be accurate, inaccurate, or irrelevant. Answer the question with discretion regarding how you use these details.\\
Description: <DESCRIPTION>\\
Question: <QUESTION>
\end{tcolorbox}
\caption{The prompt used for evaluating LALMs' response to potentially inconsistent audio-text pairs. }
    \label{fig:prompt_eval}
\end{figure}

\noindent\textbf{Accuracy.} For each textual description type, we calculate the accuracy as: $$\text{Acc}_{t} = \frac{C_{t}}{N}.$$

\noindent\textbf{Normalized Accuracy.} This metric measures how the model is affected by different types of textual input. It can be expressed as: $$\text{Norm}_{t} = \frac{C_{t}}{C_{\text{neu}}}.$$

\noindent\textbf{Macro Accuracy.} This metric is defined as the average accuracy of three different types:
$$\text{Macro} = \frac{\text{Acc}_{{\textit{fth}}} + \text{Acc}_{{\textit{adv}}} + \text{Acc}_{{\textit{irr}}}}{3}.$$

\noindent\textbf{Text Influence Rate (TIR).} TIR quantifies how much the textual input influences the model's predictions across different text conditions. We define $\Delta_{\text{ic}}^{t}$ as the number of samples with prediction changes from incorrect to correct and $\Delta_{\text{ci}}^{t}$ as those changing from correct to incorrect under text condition $t$. TIR for each text condition is:
$$
    \text{TIR}_{t} = 
    \begin{cases}
        \frac{\Delta_{\text{ic}}^{{\textit{fth}}}}{N} & \text{if } t = {\textit{fth}} \\
        \frac{\Delta_{\text{ci}}^{{\textit{adv}}}}{N} & \text{if } t = {\textit{adv}} \\
        \frac{\Delta_{\text{ic}}^{{\textit{irr}}} + \Delta_{\text{ci}}^{{\textit{irr}}}}{N} & \text{if } t = {\textit{irr}}
    \end{cases}
$$
It evaluates how the model utilizes faithful information, opposing misleading information and ignoring irrelevant information. 

\noindent\textbf{Modal Robustness Score (MRS).} This metric quantifies the resistance to potentially misleading textual information, indicating how well a model maintains audio-based performance despite contradictory or irrelevant textual inputs:
$$\text{MRS} = \alpha \cdot \frac{\text{Acc}_{\textit{adv}}}{\text{Acc}_{\textit{neu}}} + (1-\alpha) \cdot \frac{\text{Acc}_{\textit{irr}}}{\text{Acc}_{\textit{neu}}}$$
where $\alpha=0.8$ emphasizes adversarial robustness, as most models exhibit near-zero $\text{Acc}_{\textit{adv}}$ values.

\section{Benchmarking Text Bias}
Our evaluation encompasses a diverse range of state-of-the-art LALMs: Qwen-Audio-Chat~\citep{qwen-audio-chat}, Qwen2-Audio-Instruct~\citep{qwen-2-audio}, Gazelle~\citep{gazelle}, SALMONNN-7B and SALMONNN-13B~\citep{salmonn}, Audio-Flamingo2~\citep{audioflamingo2} and SeaLLMs-Audio-7B~\citep{SeaLLMs-Audio}.

\begin{table*}[ht]
\centering
{\renewcommand{\arraystretch}{1.3} 
\resizebox{1.02\textwidth}{!}{%
\begin{tabular}{lcccccccccccccc} 
\toprule
\multirow{2}{*}{\textbf{Benchmark Task}} 
& \multirow{2}{*}{\textbf{Model}} 
& \multirow{2}{*}{\textbf{Neutral}} 
& \multicolumn{3}{c}{\textbf{Faithful}} 
& \multicolumn{3}{c}{\textbf{Adversarial}} 
& \multicolumn{3}{c}{\textbf{Irrelevant}} 
& \multirow{2}{*}{\textbf{Macro}} 
& \multirow{2}{*}{\textbf{MRS}}\\
\cmidrule(lr){4-6} 
\cmidrule(lr){7-9} 
\cmidrule(lr){10-12}
 &  &  & \textbf{Accuracy $\uparrow$} & \textbf{Norm $\uparrow$} & \textbf{TIR $\uparrow$} 
 & \textbf{Accuracy $\uparrow$} & \textbf{Norm $\uparrow$} & \textbf{TIR $\downarrow$} 
 & \textbf{Accuracy $\uparrow$} & \textbf{Norm $\uparrow$} & \textbf{TIR $\downarrow$} 
 & \\
 
\midrule
\multirow{7}{*}{AQA} 
& Qwen-Audio-Chat & \shadecell{87.8} & \shadecell{100.0} & \shadecell{113.9} & \shadecell{100.0} & \shadecell{1.7} & \shadecell{1.9} & \shadecell{98.3} & \shadecell{87.9} & \shadecell{100.1} & \shadecell{12.7} & \shadecell{63.2} & \shadecell{21.5} \\ 
& Qwen2-Audio-Instruct & \shadecell{87.5} & \shadecell{100.0} & \shadecell{114.3} & \shadecell{100.0} & \shadecell{1.5} & \shadecell{1.7} & \shadecell{98.3} & \shadecell{75.5} & \shadecell{86.3} & \shadecell{27.0} & \shadecell{59.0} & \shadecell{18.6}\\
& SALMONN-7B & \shadecell{62.2} & \shadecell{99.4} & \shadecell{159.8} & \shadecell{98.7} & \shadecell{1.7} & \shadecell{2.7} & \shadecell{97.3} & \shadecell{73.8} & \shadecell{118.6} & \shadecell{26.0} & \shadecell{58.3} & \shadecell{25.9}\\
& SALMONN-13B & \shadecell{70.0} & \shadecell{99.4} & \shadecell{142.0} & \shadecell{98.3} & \shadecell{2.7} & \shadecell{3.9} & \shadecell{96.6} & \shadecell{55.3} & \shadecell{79.0} & \shadecell{62.1} & \shadecell{52.5} & \shadecell{18.9} \\
& Gazelle & \shadecell{60.5} & \shadecell{87.2} & \shadecell{144.1} & \shadecell{86.1} & \shadecell{3.5} & \shadecell{5.8} & \shadecell{96.4} & \shadecell{43.6} & \shadecell{72.1} & \shadecell{53.7} & \shadecell{44.8} & \shadecell{19.1}\\
& Audio-Flamingo2 & \shadecell{68.0} & \shadecell{90.4} & \shadecell{132.9} & \shadecell{82.5} & \shadecell{35.3} & \shadecell{51.9} & \shadecell{58.7} & \shadecell{57.5} & \shadecell{84.6} & \shadecell{38.7} & \shadecell{61.1} & \shadecell{58.4}\\
& SeaLLMs-Audio-7B & \shadecell{72.8} & \shadecell{99.9} & \shadecell{137.2} & \shadecell{99.6} & \shadecell{1.3} & \shadecell{1.8} & \shadecell{98.4} & \shadecell{81.8} & \shadecell{112.4} & \shadecell{15.6} & \shadecell{61.0} & \shadecell{23.9}\\

\midrule
\multirow{7}{*}{VSC}
& Qwen-Audio-Chat & \shadecell{60.1} & \shadecell{79.5} & \shadecell{132.3} & \shadecell{51.9} & \shadecell{3.0} & \shadecell{5.0} & \shadecell{96.7} & \shadecell{45.3} & \shadecell{75.4} & \shadecell{15.7} & \shadecell{42.6} & \shadecell{19.1} \\ 
& Qwen2-Audio-Instruct & \shadecell{85.4} & \shadecell{99.8} & \shadecell{116.9} & \shadecell{98.6} & \shadecell{11.8} & \shadecell{13.8} & \shadecell{86.2} & \shadecell{85.7} & \shadecell{100.4} & \shadecell{9.9} & \shadecell{65.8} & \shadecell{31.1} \\
& SALMONN-7B & \shadecell{60.5} & \shadecell{89.6} & \shadecell{148.1} & \shadecell{73.7} & \shadecell{25.1} & \shadecell{41.5} & \shadecell{59.0} & \shadecell{61.4} & \shadecell{101.5} & \shadecell{4.5} & \shadecell{58.7} & \shadecell{53.5} \\
& SALMONN-13B & \shadecell{48.8} & \shadecell{65.3} & \shadecell{133.8} & \shadecell{38.7} & \shadecell{24.4} & \shadecell{50.0} & \shadecell{52.5} & \shadecell{42.4} & \shadecell{86.9} & \shadecell{12.6} & \shadecell{44.0} & \shadecell{57.4} \\
& Gazelle & \shadecell{18.2} & \shadecell{100.0} & \shadecell{549.5} & \shadecell{100.0} & \shadecell{0.0} & \shadecell{0.0} & \shadecell{100.0} & \shadecell{16.9} & \shadecell{92.9} & \shadecell{15.3} & \shadecell{39.0} & \shadecell{18.6}\\
& Audio-Flamingo2 & \shadecell{30.0} & \shadecell{98.8} & \shadecell{329.3} & \shadecell{98.7} & \shadecell{1.3} & \shadecell{4.3} & \shadecell{97.3} & \shadecell{25.3} & \shadecell{84.3} & \shadecell{26.7} & \shadecell{41.8} & \shadecell{20.3} \\
& SeaLLMs-Audio-7B & \shadecell{65.2} & \shadecell{98.4} & \shadecell{150.9} & \shadecell{95.4} & \shadecell{7.1} & \shadecell{10.9} & \shadecell{88.3} & \shadecell{49.7} & \shadecell{76.2} & \shadecell{18.7} & \shadecell{51.7} & \shadecell{24.0} \\
\midrule

\multirow{7}{*}{SER} 
& Qwen-Audio-Chat & \shadecell{24.5} & \shadecell{99.9} & \shadecell{407.8} & \shadecell{99.9} & \shadecell{0.1} & \shadecell{0.4} & \shadecell{99.6} & \shadecell{14.8} & \shadecell{60.4} & \shadecell{25.9} & \shadecell{38.3} & \shadecell{12.4}\\ 
& Qwen2-Audio-Instruct & \shadecell{41.8} & \shadecell{100.0} & \shadecell{239.2} & \shadecell{100.0} & \shadecell{0.0} & \shadecell{0.0} & \shadecell{100.0} & \shadecell{27.8} & \shadecell{66.5} & \shadecell{39.4} & \shadecell{42.6} & \shadecell{13.3} \\
& SALMONN-7B & \shadecell{25.1} & \shadecell{98.7} & \shadecell{393.2} & \shadecell{98.3} & \shadecell{0.1} & \shadecell{0.4} & \shadecell{99.6} & \shadecell{36.4} & \shadecell{145.0} & \shadecell{36.1} & \shadecell{45.1} & \shadecell{29.3} \\
& SALMONN-13B & \shadecell{46.9} & \shadecell{100.0} & \shadecell{213.2} & \shadecell{100.0} & \shadecell{0.0} & \shadecell{0.0} & \shadecell{100.0} & \shadecell{45.3} & \shadecell{96.6} & \shadecell{5.0} & \shadecell{48.4} & \shadecell{19.3} \\
& Gazelle & \shadecell{44.9} & \shadecell{97.4} & \shadecell{216.9} & \shadecell{95.6} & \shadecell{0.0} & \shadecell{0.0} & \shadecell{100.0} & \shadecell{43.8} & \shadecell{97.6} & \shadecell{7.3} & \shadecell{47.1} & \shadecell{19.5} \\
& Audio-Flamingo2 & \shadecell{30.8} & \shadecell{80.2} & \shadecell{260.4} & \shadecell{76.0} & \shadecell{15.9} & \shadecell{51.6} & \shadecell{72.7} & \shadecell{32.9} & \shadecell{106.8} & \shadecell{31.1} & \shadecell{43.0} & \shadecell{62.6} \\
& SeaLLMs-Audio-7B & \shadecell{49.9} & \shadecell{99.9} & \shadecell{200.2} & \shadecell{99.8} & \shadecell{0.1} & \shadecell{0.2} & \shadecell{99.8} & \shadecell{47.2} & \shadecell{94.6} & \shadecell{18.3} & \shadecell{49.1} & \shadecell{19.1} \\

\bottomrule
\end{tabular}
}
}
\caption{\textbf{Performance comparison (\%) of various LALMs on \dataset{}}. Results show accuracy and Text Influence Rate (TIR) across neutral, faithful, adversarial, and irrelevant text inputs. Darker background color indicate higher value.}
\vspace{-10pt}
\label{tab:performance}
\end{table*}

\subsection{Main Results}
Experimental results are summarized in Table~\ref{tab:performance}. 
We observe strong text bias across all models. LALMs consistently prioritize textual information over audio evidence when faced with contradictions between modalities, regardless of their model architecture or underlying training methodology. When provided with adversarial textual descriptions that contradict audio content, all models exhibit dramatic performance drops. For instance, on the Audio Question Answering task, accuracies drops from 87.8\% to 1.7\% for Qwen-Audio-Chat and from 87.5\% to 1.5\% for Qwen2-Audio-Instruct—representing over 98\% performance deterioration. Even more strikingly, on the Speech Emotion Recognition task, four of the seven tested models show complete susceptibility to adversarial text, with accuracy dropping to precisely 0.0\%. This pattern holds across all datasets, with TIR consistently above 95\% for most models, clearly demonstrating that these systems overwhelmingly favor textual inputs when resolving cross-modal conflicts.

\subsection{Comparisons Across Models}
While text bias is universal across all tested models, some demonstrate notably higher resilience to misleading textual inputs than others. Audio-Flamingo2 stands out with substantially stronger modal robustness compared to other models, achieving significantly higher adversarial accuracy on Audio Question Answering task (35.3\% versus below 3.5\% for most competitors) and maintaining an MRS of 58.4\%. Similarly, on the Speech Emotion Recognition task, Audio-Flamingo2 maintains 15.9\% accuracy under adversarial conditions while most other models drop to near zero. SALMONN models also demonstrate relatively better resilience on Vocal Sound Classification, with the 7B and 13B versions maintaining 25.1\% and 24.4\% accuracy respectively under adversarial conditions, compared to 3.0\% of Qwen-Audio-Chat. These quantitative differences suggest meaningful variations in how different architectures integrate and prioritize cross-modal information, though even the most robust models still show considerable vulnerability to text bias.

To investigate the relationship between parameter count and cross-modal robustness, we evaluated Audio-Flamingo2~\citep{audioflamingo2}  at three different scales (0.5B, 1.5B, and 3B) as detailed in Table~\ref{tab:model_size}. Our analysis reveals a consistent performance improvement as model size increases, with the largest 3B variant showing enhanced capabilities in both leveraging helpful textual information and resisting misleading inputs. However, the relatively modest gains in adversarial resistance compared to the significant parameter increase suggest that architectural innovations, rather than simple scaling, may be necessary to effectively address cross-modal conflicts.

\subsection{Impact of Tasks and Text Relevance}

The severity of text bias varies significantly across different audio understanding tasks, revealing a relationship between task complexity and susceptibility to misleading text. LALMs show particularly high vulnerability on emotion recognition tasks, where average adversarial accuracy across all models is just 2.3\%, compared to 6.7\% on Audio Question Answering task and 10.4\% on Vocal Sound Classification task. Similarly striking is how irrelevant text affects performance differently across tasks—on Audio Question Answering, SeaLLMs-Audio-7B achieves 112.4\% normalized accuracy with irrelevant text (improved performance), while on Speech Emotion Recognition task, SALMONN-7B reaches 145.0\% of its neutral performance with irrelevant text. This variability in responses to different types of textual interference suggests that the interplay between audio and text processing is highly task-dependent, with semantically complex tasks showing different vulnerability patterns than more straightforward classification tasks.

\begin{figure}[h]
   \centering
   \includegraphics[width=0.9\linewidth]{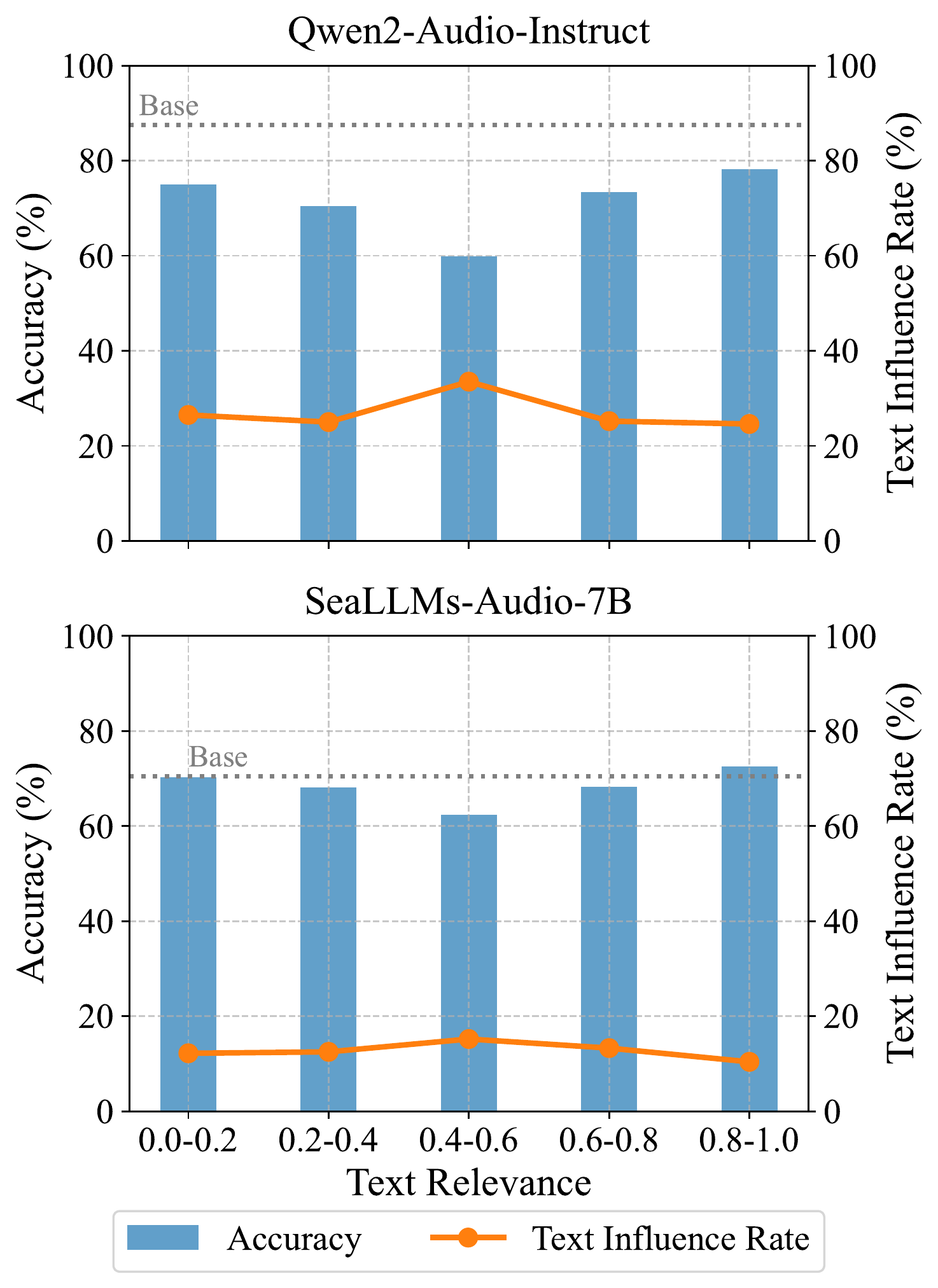}
   \caption{\textbf{Analysis of Text Relevance Impact.} Performance across five text relevance bins from lowest to highest. Blue bars (left axis) show accuracy under adversarial text conditions, while the orange line (right axis) represents the Text Influence Rate.}
   \label{fig:relevance}
\end{figure}

To investigate how textual relevance affects model behavior, we quantify the semantic distance between textual descriptions and audio content, dividing samples into five bins from lowest to highest relevance. Using sentence embeddings to compute cosine similarity between text and audio captions, we evaluate performance across these relevance levels. As shown in Figure~\ref{fig:relevance}, surprisingly, there is no clear correlation between text relevance and the model's susceptibility to textual bias. The Text Influence Rate remains consistently high across all relevance bins for adversarial text, suggesting that LALMs' text bias persists regardless of semantic distance between modalities.

\begin{table}[t]
    \centering
    \footnotesize
    \begin{tabular}{lccccc}
        \toprule
        \multirow{2.5}{*}{\textbf{Size}} & \multicolumn{3}{c}{\textbf{Text Influence Rate}} & \multirow{2.5}{*}{\textbf{Macro} ↑} & \multirow{2.5}{*}{\textbf{MRS} ↑} \\
        \cmidrule(lr){2-4}
        & Faith. ↑ & Adv. ↓ & Irr. ↓ & & \\
        \midrule
        0.5B & 75.36 & 66.38 & 44.20 & 54.60 & 58.10 \\
        1.5B & 72.67 & 59.30 & 43.80 & 55.17 & 59.44 \\
        3B & \textbf{82.50} & \textbf{58.68} & \textbf{38.70} & \textbf{61.07} & \textbf{60.68} \\
        \bottomrule
    \end{tabular}
    \caption{\textbf{The Effect of Model Sizes.} We experiment with Audio-Flamingo2 at three different parameter scales on ClothoAQA.}
    \vspace{-10pt}
    \label{tab:model_size}
\end{table}

\section{Understanding Text Bias}

We perform in-depth analysis to disclose the causes of text bias in LALMs. 

\subsection{Confidence Analysis}
To investigate whether LALMs exhibit appropriate uncertainty when faced with inconsistent inputs, we analyze confidence patterns in Qwen2-Audio-Instruct and SeaLLMs-Audio-7B across different textual conditions. For each prediction, we extract the maximum token probability as a confidence score, allowing us to quantify model certainty under modal conflict.

\begin{figure}[h]
    \centering   \includegraphics[width=0.95\linewidth]{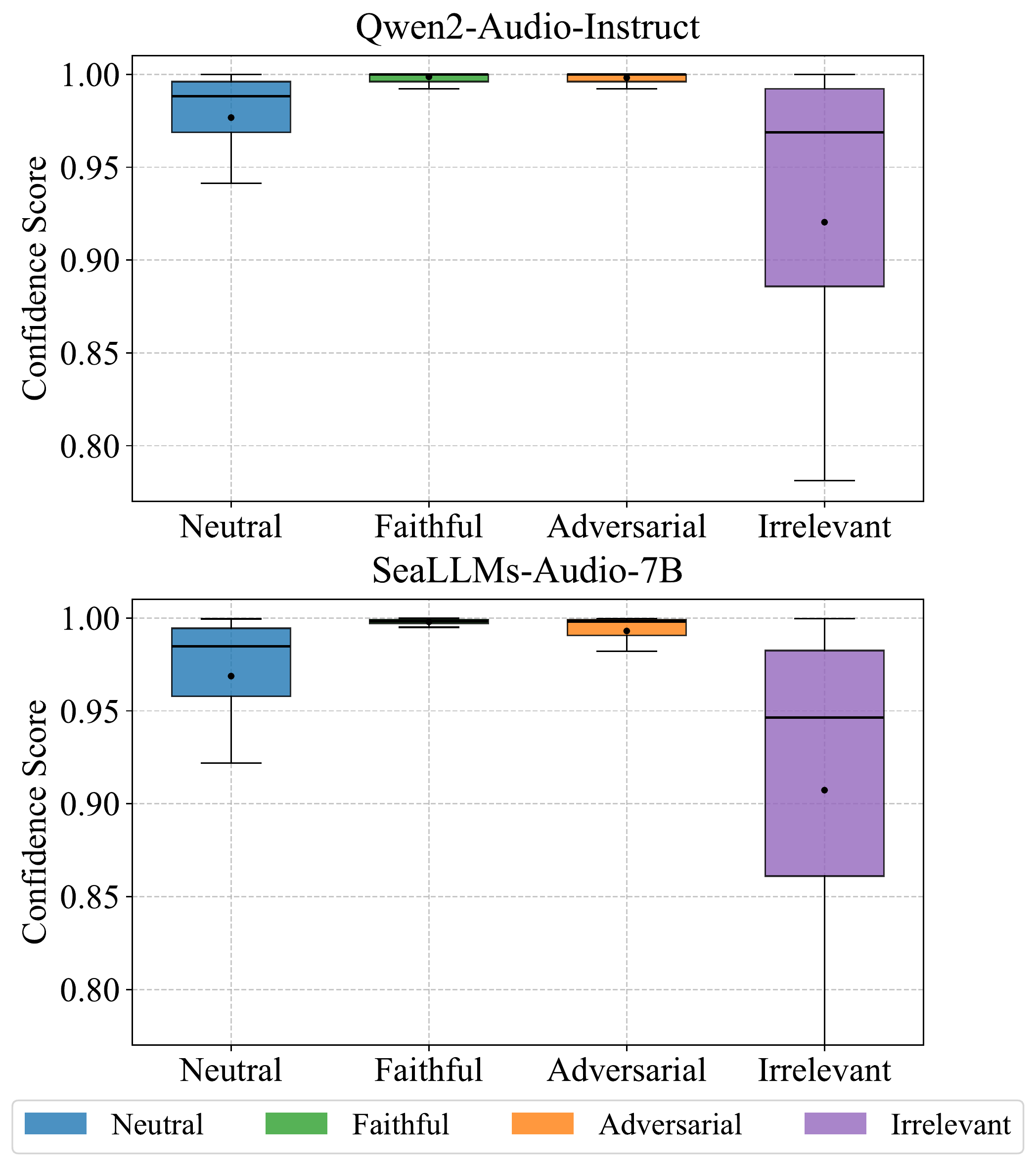}
    \caption{\textbf{Confidence Analysis Under Different Textual Conditions.} LALMs maintain high confidence scores across text conditions despite performance degradation with adversarial inputs.}
    \label{fig:confidence}
\end{figure}

As shown in Figure~\ref{fig:confidence}, LALMs maintain remarkably high confidence scores even when processing adversarial textual inputs that contradict audio evidence. Surprisingly, confidence under adversarial conditions is comparable to or even higher than under faithful conditions, despite the dramatic performance degradation observed in our earlier experiments. Only with irrelevant text do we observe a slight reduction in confidence, though this decrease remains disproportionately small relative to performance impact. This overconfidence when making incorrect predictions indicates that LALMs not only prioritize text over audio but also do so with high certainty, suggesting these models lack effective calibration mechanisms to detect and appropriately respond to cross-modal inconsistencies.

\subsection{Spectral Analysis}
We analyze the intrinsic dimensionality of hidden representations when processing consistent versus inconsistent audio-text pairs. For $N$ samples, we extract the last layer hidden states of the final token, resulting in two matrices: $A \in \mathbb{R}^{N \times d}$ from adversarial inputs and $F \in \mathbb{R}^{N \times d}$ from faithful inputs, where $d$ represents the hidden state dimension. After centralizing these matrices, we perform Singular Value Decomposition (SVD):
$$
A = U_A \Sigma_A V_A^T, \quad F = U_F \Sigma_F V_F^T
$$
where $U_A, U_F \in \mathbb{R}^{N \times N}$ and $V_A, V_F \in \mathbb{R}^{d \times d}$ are orthogonal matrices.

\begin{figure}[t]
    \centering
    \includegraphics[width=0.8\linewidth]{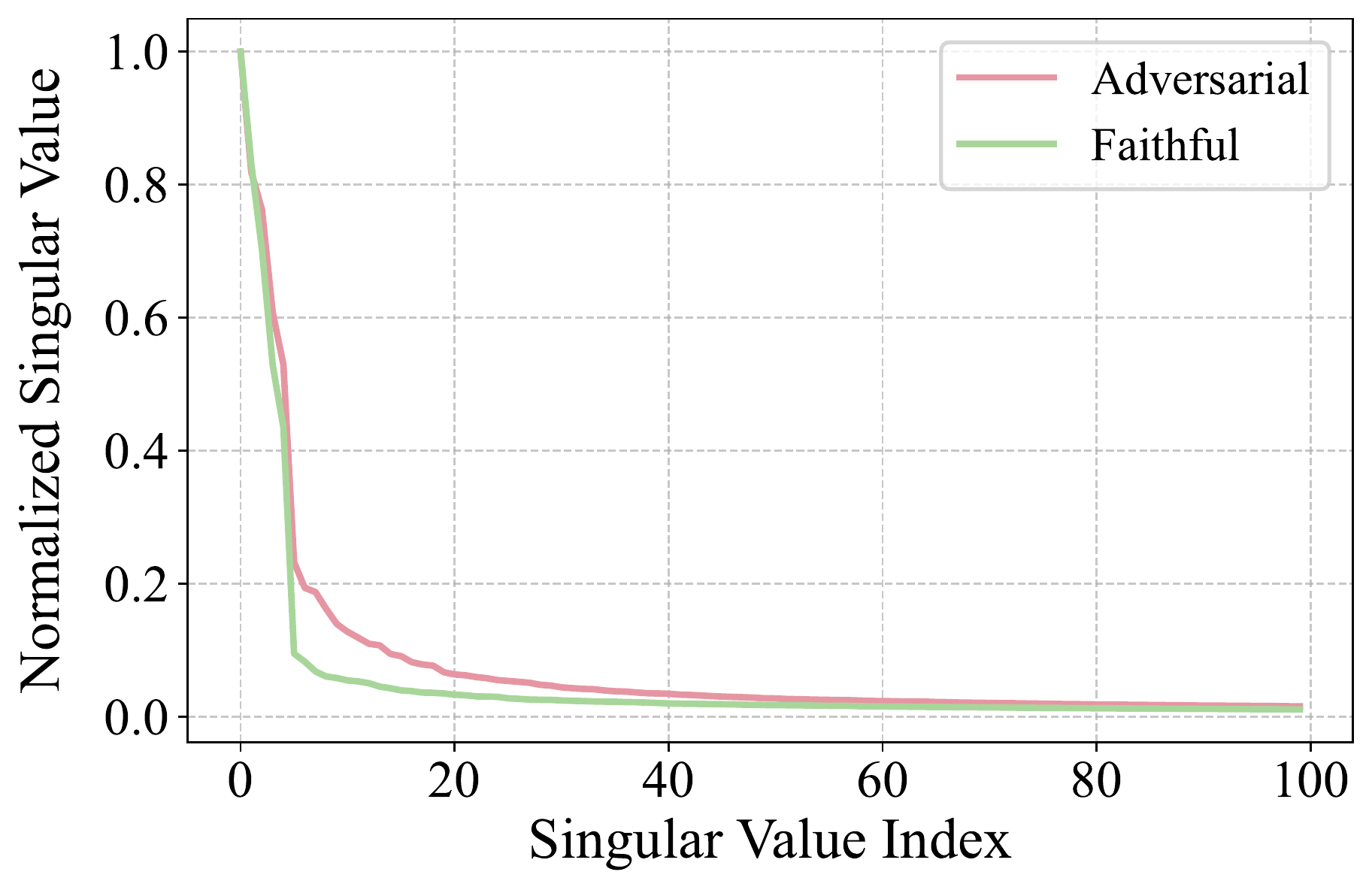}
    \caption{\textbf{Spectral analysis of hidden representations.} Normalized singular values for adversarial and faithful inputs from Qwen2-Audio-Instruct on subset of \dataset{}.}
    \vspace{-15pt}
    \label{fig:svd}
\end{figure}

Using Qwen2-Audio-Instruct with the Vocal Sound Classification subset of \dataset{}, we plot the normalized singular values in Figure~\ref{fig:svd}. The results reveal a rapid decay in singular values for both conditions, indicating that the model's representations lie in remarkably low-dimensional subspaces. Specifically, only 6 dimensions are needed to explain 95\% of the variance in adversarial representations, while faithful representations require just 5 dimensions. This suggests that despite the high-dimensional embedding space, the model encodes audio-text information in compact, low-dimensional manifolds.

\subsection{Separability Analysis}
Building on our spectral analysis findings, we further investigate the separability between these low-dimensional subspaces. If the model internally distinguishes between faithful and adversarial inputs—despite producing confident yet incorrect outputs for adversarial cases—these subspaces should be linearly separable. We implement a 3:1 train-test split on the hidden representations from different model layers and train SVM and Random Forest classifiers to quantify this separability.

Table~\ref{tab:hidden_states} presents the classification performance across different layers. The high accuracy (up to 98.0\% with Random Forest at layer 32) confirms that these representation subspaces are highly separable, with the separation becoming more pronounced in deeper layers. This indicates that LALMs internally recognize inconsistencies between audio and text modalities, yet this awareness fails to translate into appropriate output behavior—revealing a disconnect between representation and decision-making in these models.

\begin{table}[h]
\small
\renewcommand{\arraystretch}{1.0}
\centering
\vspace{-5pt}
\begin{tabular}{lcccc}
\toprule
\textbf{Method} & \textbf{Layer} & \textbf{Acc} & \textbf{F1} & \textbf{AUC} \\
\midrule
\multirow{3}{*}{SVM} & 1 & 48.2 & 51.0 & 53.8\\
& 16 & 93.4 & 93.6 & 97.9 \\
& 32 & 95.8 & 95.9 & 98.8 \\
\midrule
\multirow{3}{*}{Random Forest} & 1 & 56.4 & 58.6 & 60.4 \\
& 16 & 97.4 & 97.4 & 99.5\\
& 32 & 98.0 & 98.0 & 99.8\\
\bottomrule
\end{tabular}
\caption{\textbf{Subspace Classification Performance.} We train SVM and Random Forest Classifier on the adversarial and faithful input.}
\label{tab:hidden_states}
\end{table}

\section{Mitigating Text Bias}

We discuss two potential solutions to mitigate the text bias in LALMs. 


\subsection{Prompting Techniques}
Inspired by previous studies~\citep{shi2023large, ailindeng}, we first investigate whether different prompting techniques will help models reduce the text bias. We consider the following techniques: Zero-Shot Chain-of-Thought prompting~\citep{kojima2023largelanguagemodelszeroshot}, Audio Priority prompting which explicitly instructs the model to prioritize audio information, and Bias Awareness prompting which reminds the model about potential modality conflicts (prompts are shown in Appendix~\ref{app:mitigation-prompts}). 

\begin{figure}[t]
    \centering
    \includegraphics[width=1.04\linewidth]{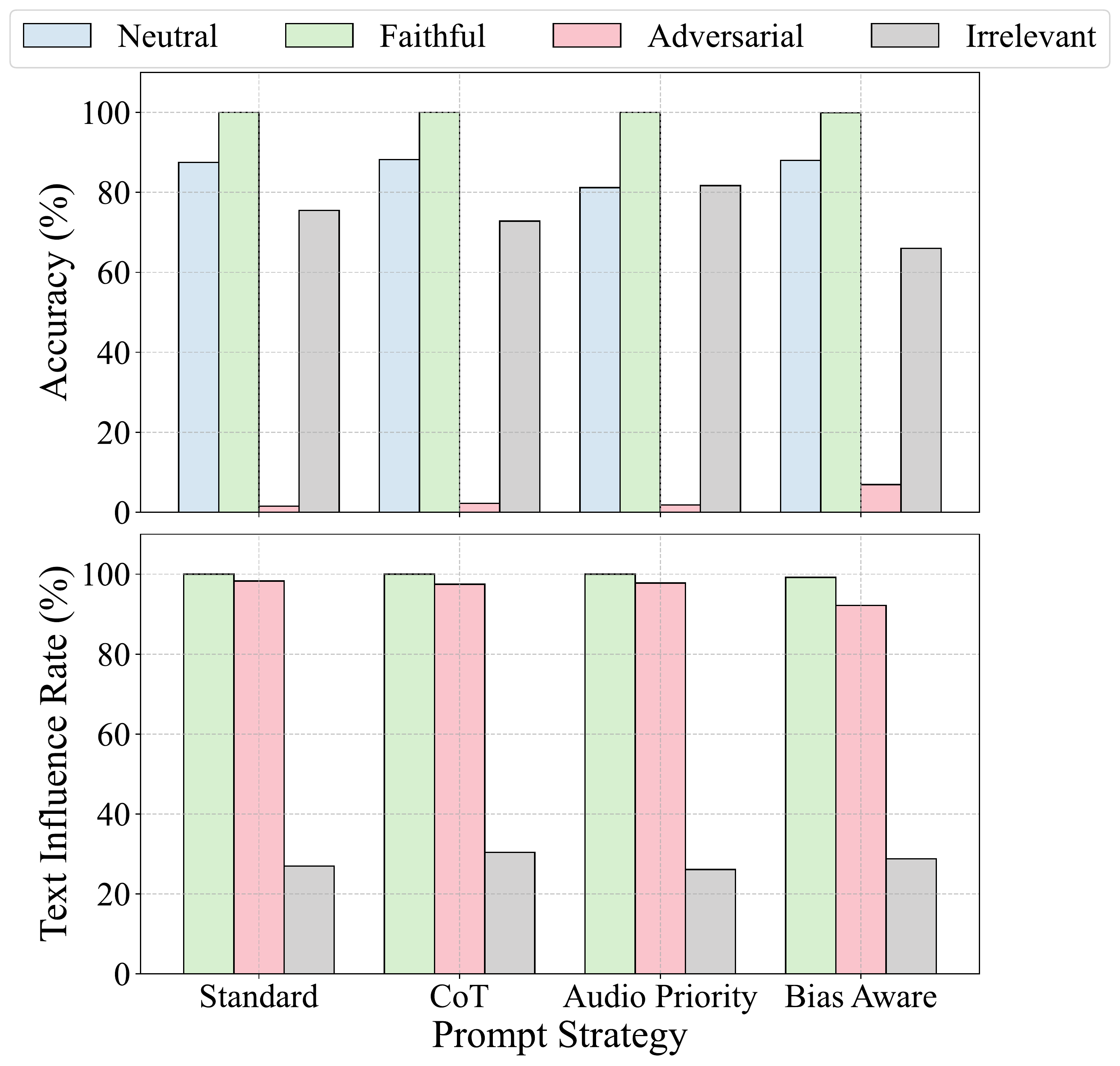}
    \caption{\textbf{Analysis on Different Prompting Techniques.} We perform our experiments on Qwen2-Audio-Instruct and \dataset{} subset.}
    \vspace{-10pt}
    \label{fig:prompt}
\end{figure}

\begin{table*}[t]
\centering
\small
\setlength{\tabcolsep}{4pt}
\renewcommand{\arraystretch}{1.25} 
\begin{tabular}{lcccccccccccc}
\toprule
\multirow{2}{*}{\textbf{Method}} & \multicolumn{4}{c}{\textbf{ClothoAQA}} & \multicolumn{4}{c}{\textbf{MELD}} & \multicolumn{4}{c}{\textbf{VocalSound (Out-of-Distribution)}} \\
\cmidrule(lr){2-5} \cmidrule(lr){6-9} \cmidrule(lr){10-13}
& Acc$_\textit{fth}$ & Acc$_\textit{adv}$ & Acc$_\textit{irr}$ & TIR$_\textit{adv}$ 
& Acc$_\textit{fth}$ & Acc$_\textit{adv}$ & Acc$_\textit{irr}$ & TIR$_\textit{adv}$ 
& Acc$_\textit{fth}$ & Acc$_\textit{adv}$ & Acc$_\textit{irr}$ & TIR$_\textit{adv}$ \\
\midrule
Base & 100.0 & 1.5 & 75.5 & 98.3 & 100.0 & 0.0 & 27.8 & 100.0 & 99.8 & 11.8 & 85.7 & 86.2 \\
\midrule
\multicolumn{13}{c}{\textit{Prompt-based Methods}} \\
\midrule
w/ CoT & 100.0 & 2.2 & 72.8 & 97.5 & 100.0 & 0.0 & 28.5 & 100.0 & 99.8 & 11.9 & 84.9 & 86.1 \\
w/ Bias Awareness & 99.9 & 6.9 & 66.0 & 92.2 & 100.0 & 0.0 & 28.0 & 100.0 & 100.0 & 11.8 & 85.1 & 86.3 \\
w/ Audio Priority & 100.0 & 1.8 & 81.7 & 97.8 & 100.0 & 0.0 & 26.2 & 100.0 & 99.9 & 12.1 & 85.2 & 86.0 \\
\rowcolor[gray]{0.95} Best Prompt & 100.0 & 6.9 & 81.7 & 92.2 & 100.0 & 0.0 & 28.5 & 100.0 & 10.0 & 12.1 & 85.7 & 86.3 \\
\midrule
\multicolumn{13}{c}{\textit{Finetuning-based Methods}} \\
\midrule
w/ SFT & 90.9 & 42.1 & 89.2 & 18.7 & 60.6 & 43.8 & 47.2 & 14.6 & 96.2 & 17.7 & 92.1 & 76.9 \\
\bottomrule
\end{tabular}
\caption{\textbf{Comparison of techniques for mitigating text bias in Qwen2-Audio-Instruct.} We compare the base model, bias awareness prompting, and SFT across training datasets and evaluate generalization on the out-of-distribution subset.}
\label{tab:SFT}
\vspace{-15pt}
\end{table*}

In our experiments with Qwen2-Audio-Instruct on the Audio Question Answering subset of \dataset{} (result in Figure~\ref{fig:prompt}), we find that prompting techniques alleviate the text bias to some extent, but the improvement is very limited.
Specifically, the Bias Awareness prompt shows the most significant effect, increasing the accuracy from 1.5\% to 17.4\% under adversarial conditions. It also decreases the Text Influence Rate from 98.3\% to 79.7\%, indicating reduced susceptibility to misleading text. However, even with these improvements, the model's performance remains compromised when faced with contradictory textual information, suggesting that more fundamental architectural or training modifications may be necessary to effectively address the text bias problem  in LALMs.

\subsection{Supervised Finetuning (SFT)}
We investigate whether supervised finetuning (SFT) on datasets containing conflicting audio-text pairs can mitigate text bias in LALMs. This strategy explicitly trains the model to recognize and resolve cross-modal inconsistencies by providing the correct answers despite misleading textual information. This targeted intervention aims to recalibrate the model's attention between modalities when faced with conflicting inputs. 

We use Qwen2-Audio-Instruct as our base model, and fine-tune it on 1,000 samples from Audio Question Answering and Speech Emotion Recognition subsets that contain deliberately mismatched audio-text pairs. To ensure efficient adaptation while preserving general capabilities, we employ Low-Rank Adaptation (LoRA)~\citep{hu2022lora} with a rank of 8 and train for 2 epochs. Fine-tuning details are given in Appendix~\ref{app:SFT}. We evaluate the model's generalization on the Vocal Sound Classification task, which represents an unseen domain.

Table~\ref{tab:SFT} presents the performance of the base and fine-tuned models across different metrics. We observe that SFT substantially outperforms prompt-based methods in mitigating text bias. Our fine-tuned model shows dramatically improved adversarial accuracy across all datasets, with particularly notable gains on Audio Question Answering and Speech Emotion Recognition tasks. This comes with a significant reduction in Text Influence Rate, indicating enhanced resistance to misleading textual cues. However, this improvement trades off some performance on faithful text conditions, suggesting a recalibration of modality attention rather than an overall enhancement. Interestingly, the model exhibits improved handling of irrelevant textual inputs as well, demonstrating more balanced cross-modal processing. Despite these gains, text bias remains present, highlighting the need for more advanced architectural approaches to fully resolve modality imbalance in LALMs.

\section{Conclusion}

In this work, we introduce \dataset{}, a benchmark that evaluates the performance of LALMs when faced with cross-modal inconsistencies. Our comprehensive evaluations across multiple models and tasks demonstrate that state-of-the-art LALMs exhibit a strong bias towards textual input over audio, leading to consistent performance degradation under adversarial conditions. 
We explore various mitigation strategies, which can only partially address the issue. 
These findings highlight the critical reliability concerns for real-world applications and underscore the need for novel training paradigms to better balance modality contributions in multi-modal processing. We believe \dataset{} will serve as a valuable benchmark for developing more robust large audio-language models.

\section*{Limitations}

Despite our comprehensive evaluation, this study has several limitations. Our analysis is constrained to specific audio understanding tasks and may not generalize to all audio-language scenarios. The synthetic nature of our adversarial and irrelevant textual descriptions might present different challenges compared to naturally occurring conflicts. Our investigation of mitigation strategies was limited to prompting techniques and model scaling, without exploring architectural modifications or specialized training objectives that could potentially yield more substantial improvements. Additionally, our evaluation focused on English-language models and Western audio contexts, potentially missing cultural and linguistic factors that may influence cross-modal processing priorities.

\section*{Acknowledgments}

This research is supported by the National Research Foundation, Singapore and Infocomm Media Development Authority under its Trust Tech Funding Initiative. Any opinions, findings and conclusions or recommendations expressed in this material are those of the author(s) and do not reflect the views of National Research Foundation, Singapore and Infocomm Media Development Authority. This work is supported by the National Research Foundation, Singapore, and Cyber Security Agency of Singapore under its National Cybersecurity R\&D Programme and CyberSG R\&D Cyber Research Programme Office. Any opinions, findings and conclusions or recommendations expressed in these materials are those of the author(s) and do not reflect the views of National Research Foundation, Singapore, Cyber Security Agency of Singapore as well as CyberSG R\&D Programme Office, Singapore.

\bibliographystyle{acl_natbib}
\bibliography{custom}

\vfill
\appendix

\section{Faithful and Adversarial Statement Generation}
\label{app:faithful-adv-prompt}
The prompt used to generate faithful statements that accurately reflect audio content and adversarial statements that contradict the audio content is presented in Figure~\ref{fig:prompt_annotate}.
\begin{figure}[H]
    \centering
\begin{tcolorbox}[fonttitle=\bfseries,title=Prompt for Text Variants Generation, size=normal, label=mybox]
Convert this question and answer into two statements:\\
1. A factual statement that accurately represents the information from the question and answer.\\
2. A non-factual statement that contradicts the factual statement.\\

\textbf{Example:}\\
Question: "Are people speaking?"\\
Answer: "yes"\\
Factual statement: "There are people speaking."\\
Non-factual statement: "There are no people speaking."\\

\textbf{Now convert this pair:}\\
Question: "<QUESTION>"\\
Answer: "<ANSWER>"\\
Factual statement:\\
Non-factual statement:
\end{tcolorbox}
\caption{Prompt used for generating text variants from question-answer pairs.}    
\label{fig:prompt_annotate}
\end{figure}

\section{Mitigation Strategy Prompts}
\label{app:mitigation-prompts}

This section details the prompt used for mitigating text bias of LALMs (shown in Figure~\ref{fig:prompt_cot} and Figure~\ref{fig:prompt_AP}). 

\begin{figure}[H]
    \centering
\begin{tcolorbox}[fonttitle=\bfseries,title=Audio Priority Prompt, size=normal]
After thinking step by step, please answer the question.
\end{tcolorbox}
\caption{CoT prompt for mitigating text bias.}    
\label{fig:prompt_cot}
\end{figure}

\begin{figure}[H]
    \centering
\begin{tcolorbox}[fonttitle=\bfseries,title=Audio Priority Prompt, size=normal]
Please prioritize information from the audio over the text description. 
\end{tcolorbox}
\caption{Audio Priority prompt for mitigating text bias.}    
\label{fig:prompt_AP}
\end{figure}

\begin{figure}[H]
    \centering
\begin{tcolorbox}[fonttitle=\bfseries,title=Bias Awareness Prompt, size=normal]
Be aware that you may have a tendency to trust text descriptions more than audio evidence. Try to avoid this text bias and then answer the question.
\end{tcolorbox}
\caption{Bias Awareness prompt for mitigating text bias.}    
\label{fig:prompt_bias}
\end{figure}

\section{SFT Details}
\label{app:SFT}
We fine-tuned the Qwen2-Audio-7B-Instruct model using LoRA with rank 8 and $\alpha=32$, targeting all linear layers while freezing the ViT components. Training ran for 2 epochs with a learning rate of 1e-4 and warmup ratio of 0.05. We used a per-device batch size of 1 with gradient accumulation steps of 16, resulting in an effective batch size of 128. All training was performed using bfloat16 precision with a maximum sequence length of 2048 tokens.

\end{document}